\documentclass{article} 
\usepackage{iclr2021_conference,times}

\usepackage{mathtools} 
\usepackage{booktabs} 
\usepackage[utf8]{inputenc} 
\usepackage[T1]{fontenc}    
\usepackage{url}            
\usepackage{amsfonts}       
\usepackage{nicefrac}       
\usepackage{microtype}      
\usepackage{lipsum}
\usepackage{graphicx}
\usepackage{xcolor}
\usepackage{subcaption}
\usepackage{tikz}
\usetikzlibrary{bayesnet}
\usepackage{centernot}
\usepackage[]{hyperref}
\include{commands}
\DeclareMathAlphabet{\mathcal}{OMS}{cmsy}{m}{n}

\usepackage{microtype}
\usepackage{graphicx}
\usepackage{booktabs} 
\usepackage{multirow}
\usepackage{amssymb}
\usepackage{amsmath}
\usepackage{amsthm}

\usepackage{wrapfig,lipsum,booktabs}

\usepackage{xcolor}

\def\*#1{\boldsymbol{\mathbf{#1}}}

\usepackage{hyperref}

\usepackage{amsmath,amsfonts,bm}

\def\eqref#1{equation~\ref{#1}}

\def\1{\bm{1}}

\DeclareMathAlphabet{\mathsfit}{\encodingdefault}{\sfdefault}{m}{sl}
\SetMathAlphabet{\mathsfit}{bold}{\encodingdefault}{\sfdefault}{bx}{n}

\DeclareMathOperator*{\argmin}{arg\,min}

\title{Do Concept Bottleneck Models Learn\\ As Intended?}

\author{%
  Andrei Margeloiu$^{*}$ \\
  University of Cambridge \\
  \texttt{am2770@cam.ac.uk} \\
  \And
  Matthew Ashman$^{*}$ \\
  University of Cambridge\\
 \texttt{mca39@cam.ac.uk} \\
  \And
  Umang Bhatt\thanks{Equal Contribution} \\
  University of Cambridge \\
  \texttt{usb20@cam.ac.uk} \\
  \And
  Yanzhi Chen \\
  University of Edinburgh \\
  \And
  Mateja Jamnik \\
  University of Cambridge\\
  \And
  Adrian Weller \\
  University of Cambridge\\The Alan Turing Institute \\
}

\iclrfinalcopy 
\begin{document}

\maketitle

\begin{abstract}
Concept bottleneck models map from raw inputs to concepts, and then from concepts to targets. Such models aim to incorporate pre-specified, high-level concepts into the learning procedure, and have been motivated to meet three desiderata: interpretability, predictability, and intervenability. 
However, we find that concept bottleneck models struggle to meet these goals. Using post hoc interpretability methods, we demonstrate that concepts do not correspond to anything semantically meaningful in input space, thus calling into question the usefulness of concept bottleneck models in their current form.
\end{abstract}

\section{Introduction}
\label{intro}
\citet{koh2020concept} proposed concept bottleneck models (CBMs)
as a way to incorporate pre-defined expert concepts (e.g., ``bone spurs present'' or ``wing color'') into a supervised learning procedure. 
The approach maps raw inputs ($\*x$) to concepts ($\*c$), and then maps concepts ($\*c$) to targets ($\*y$). This can be done by specifying an intermediate layer of a neural network model and aligning the layer's activations with concepts when training. As per~\citet{koh2020concept}, a CBM addresses three desiderata:
\begin{enumerate}
    \item Interpretability: Being able to note which concepts are important for the targets.
    \item Predictability: Being able to predict the targets from the concepts alone.
    \item Intervenability: Being able to replace predicted concept values with ground truth values to improve predictive performance.
\end{enumerate}

While CBMs are intuitive and have been shown in some settings to provide comparable performance to standard supervised learning approaches (directly learning a mapping between $\*x$ and $\*y$), it may often be unrealistic to assume that we will have access to concepts which fully capture the relationship between the inputs and targets. Forcing the high-dimensional inputs to pass through a low-dimension concept bottleneck might lose additional information from the inputs that is not captured by the expert-defined concepts.
Nonetheless, \citet{koh2020concept} propose three methods for training a CBM: training the input-to-concept mapping and the concept-to-target mapping \textit{jointly}; training the mappings \textit{independently}, using the ground truth concepts for the concept-to-target mapping; and training the mappings \textit{sequentially}, using the predicted concepts for the concept-to-target mapping.
While \citet{koh2020concept} favor the joint training objective, we call into question its utility for meeting the three desiderata of a CBM.
Our results suggest that training a CBM independently may be the only way the three desiderata can be achieved. 
When we explore how to visualize concepts (as we may want to show domain experts what a concept represents in input space), we find that concepts appear not to correspond to anything semantically meaningful.
We highlight our contributions:
\begin{enumerate}
    \item We investigate training a CBM using constrained bottlenecks, demonstrating that neither the joint nor the sequential training methods satisfy the three desiderata.
    \item To analyze the behavior of a CBM, we leverage post hoc interpretability techniques to understand where in input space the concepts lie.
    We find that pre-specified concepts during training do not appear to correspond to anything meaningful in input space.
\end{enumerate}

\section{Problem Setup}
\label{extension}
Consider a setting where we are given a set of inputs $\*x \in \mathbb{R}^d$ and corresponding targets $\*y \in \mathcal{Y}$. Suppose we are also given a set of pre-specified concepts $\*c \in \mathbb{R}^k$ , such that the training set comprises $\{\*x_n, y_n, \*c_n\}_{n=1}^N$. A CBM is of the form $f(g(\*x))$, where $g: \mathbb{R}^d \rightarrow \mathbb{R}^k$ maps from input space to concept space and $f: \mathbb{R}^k \rightarrow \mathcal{Y}$ maps from concept space to target space. Let $\mathcal{L}_t : \mathcal{Y} \times \mathcal{Y} \rightarrow \mathbb{R}$ denote a loss function that measures the discrepancy between the predicted and true targets and $\mathcal{L}_c : \mathbb{R}^k\times\mathbb{R}^k\rightarrow\mathbb{R}$ denote a loss function that measures the discrepancy between the predicted and true concepts. \citet{koh2020concept} propose three distinct methods for learning the functions $g$ and $f$:
\begin{enumerate}
    \item The \textbf{independent} CBM learns the mappings $g_\text{ind} = \argmin_g \sum_{n=1}^N \mathcal{L}_c(g(\*x_n); c_{n})$ and $f_\text{ind} = \argmin_f \sum_{n=1}^N \mathcal{L}_t(f(\*c_n); y_{n})$. That is, we learn a mapping from inputs to ground-truth concepts, and then separately learn a mapping from ground-truth concepts to targets.
    
    
    \item The \textbf{sequential} CBM learns $g$ in an identical manner as the independent case ($g_\text{seq} = g_\text{ind}$), but learns $f$ using the \emph{predicted} concept values from the learnt $g$, rather than using the ground-truth concept values: $f_\text{seq} = \argmin_f \sum_{n=1}^N \mathcal{L}_t(f(g_\text{seq}(\*x_n)); y_{n})$.
    \item The \textbf{joint} CBM learns $g$ and $f$ jointly by minimizing the combined objective: $f_\text{jnt}, g_\text{jnt} = \argmin_{f, g} \sum_{n=1}^N \mathcal{L}_t(f(g(\*x_n)); y_n) + \lambda \mathcal{L}_c(g(\*x_n); c_{n})$
    where $\lambda$ trades off the two losses.
\end{enumerate}
In all cases, target predictions at test time are made using the combined mapping $f(g(\*x_{\text{test}}))$. We leverage post hoc interpretability methods to identify (1) which input features are relevant to each concept and (2) which concepts are relevant to each target. We use Integrated Gradients (IG)~\citep{sundararajan2017axiomatic} to do this analysis, as IG is popular amongst practitioners~\citep{bhatt2020explainable} and performs well on multiple explanation evaluation criteria~\citep{ancona2018towards}. When visualizing saliency maps, we use Gradients~\citep{baehrens2010explain} and SmoothGrad~\citep{smilkov2017smoothgrad} too.

We consider two applications of CBMs: x-ray grading and bird identification. For x-ray grading, we use a dataset from the Osteoarthritis Initiative (OAI) with $10$ pre-specified expert concepts that correspond to clinically relevant factors~\citep{peterfy2008osteoarthritis}. For bird identification, we use the CUB dataset with $112$ concepts that capture the semantic attributes of each image~\citep{wah2011caltech}. 
These are the same two datasets considered by \citet{koh2020concept}. 
See Appendix~\ref{setup} for additional details.

\section{Checking the Three Desiderata}
\label{sec:sanity_checks}
We expect that decreasing the number of concepts in a CBM should lead to a significant drop in predictive performance, since a single concept alone may not be highly predictive of the target. To check the desiderata, we train models where we only use one concept at a time in our concept layer, $k=1$. We compare this one-concept CBM to the predicted performance of a concept oracle model that uses only the single ground truth concept to predict the targets. While the independent CBM and the concept oracle share the same concept-target predictor $f_\text{ind}$, the independent CBM obtains targets via $f_\text{ind}(g_\text{ind}(\*x_{\text{test}}))$ and the concept oracle obtains targets via $f_\text{ind}(\*c_{\text{test}})$.



Table~\ref{tab:single} demonstrates that the performance of the joint CBM far exceeds that of the concept oracle, as the joint achieves much lower error. 
This suggests that the joint CBM is learning useful information for predicting the target beyond simply getting the concept correct, that is, there is no concept bottleneck. Note that the joint CBM will be sensitive to the choice of $\lambda$; however, we use the same $\lambda$ as \citet{koh2020concept} in our experiments. 
In contrast, the independent and sequential CBMs have no incentive to learn the targets before the concept layer, and thus achieve similar predictive performance to the concept oracle. 
On reflection, this is not surprising: each concept is represented as a scalar at the concept layer, but is often categorical or binary in the real-world. Thus, extra information about the targets is likely learned in the joint CBM. When all concepts are used (rightmost column of Table~\ref{tab:single}), the concept oracle  achieves low error, suggesting that all 10 concepts together capture the target well. However, it may be difficult to obtain a full set of concepts that well specify the target.

\begin{table*}[t]
    \centering
    \small
        \caption{\small Root mean square error (RMSE) of $y$ on OAI. We report the single concept predictive performance for each of ten concepts. Results are averaged over 5 random initializations. All standard deviations were $< 0.01$.}
    \begin{tabular}{rccccccccccc}
    \toprule
    & \multicolumn{10}{c}{Concept \#} \\
    \cmidrule{2-11}
    \textbf{Model} & 1 & 2 & 3 & 4 & 5 & 6 & 7 & 8 & 9 & 10 & All concepts \\
    \midrule
    Joint & \textbf{0.49} & \textbf{0.46} & \textbf{0.46} & \textbf{0.48} & \textbf{0.47} & \textbf{0.50} & \textbf{0.47} & \textbf{0.45} & \textbf{0.48} & \textbf{0.44} & 0.42 \\
    Sequential & 0.56 & 0.56 & 0.59 & 0.57 & 0.59 & 0.61 & 0.76 & 0.75 & 0.67 & 0.75 & 0.42 \\
    Independent & 0.58 & 0.64 & 0.65 & 0.58 & 0.63 & 0.65 & 0.79 & 0.79 & 0.68 & 0.79 & 0.44 \\
    \midrule
    \small{Concept Oracle} & 0.62 & 0.64 & 0.59 & 0.59 & 0.63 & 0.70 & 0.80 & 0.78 & 0.72 & 0.80 & \textbf{0.16} \\
    \bottomrule
    \end{tabular}
    \label{tab:single}
\end{table*}

\begin{wrapfigure}{r}{7cm}
    \centering
    \includegraphics[width=\linewidth]{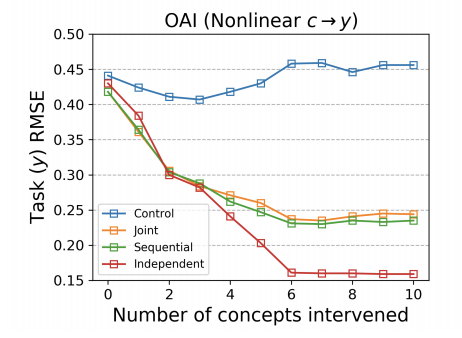}
    \caption{Intervenability on joint and sequential CBMs is better than that of the independent CBM: this is from Figure 4 of~\citep{koh2020concept}.}
    \label{fig:koh}
\end{wrapfigure}

We posit that the dependence between targets and concepts in CBMs should mirror that of the concept oracle. 
To test our hypothesis, we measure the coefficient of determination $R^2$ between saliency maps from the concept oracle ($f_{o} = f_\text{ind}(\*c_{\text{test}})$) and saliency maps (via IG) from $f_{jnt} = f_\text{jnt}(g_\text{jnt}(\*x_{\text{test}}))$ and from $f_{seq} = f_\text{seq}(g_\text{seq}(\*x_{\text{test}}))$. 
For CUB, we find that $R^2(f_{o}, f_{seq}) = 0.47 \pm 0.04$, $R^2(f_{o}, f_{jnt}) = 0.24 \pm 0.07$, $R ^2(f_{seq}, f_{jnt}) = 0.07 \pm 0.04$.
Note that high correlation does not hold between the joint or sequential CBM and the concept oracle. This agrees with our observation that the predicted concepts are \textbf{not} used in their intended manner but rather as proxies to incorporate target information.

Figure~\ref{fig:koh} indicates that, since the intervenability of the joint and sequential CBMs is inferior to that of the independent CBM as per~\citet{koh2020concept}, the concept may not be used as intended. This is likely an artifact of one-hot encoding each concept values. We suggest that future work studies how the success of intervenability varies for different concept representations: binary, scalar, one-hot encoded categoricals, etc.

We have called into question CBMs trained using the joint objective, as they seem to fall short on all three of our desiderata: (1) they do not provide concept interpretability: post hoc analysis shows that the importance of individual concepts does not correspond to their true importance in predicting the targets; (2) they do not always predict target values based on concepts, thus violating predictability; and (3) they may not intervenable, as the concepts are learned at concept layer. 

\section{Concepts in Input Space}
\begin{figure*}[htb]
    \centering
    \includegraphics[width=0.7\textwidth]{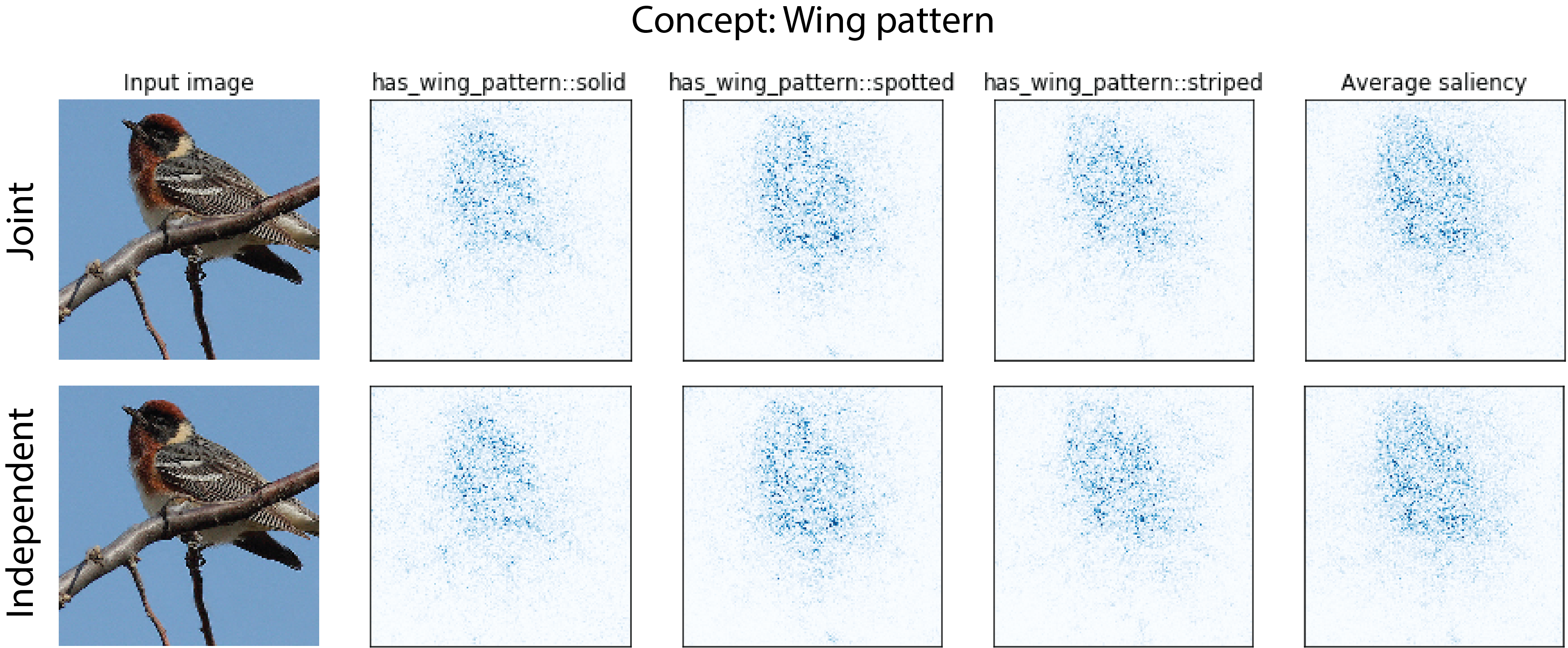}
    \caption{\small Post hoc comparison between the joint and independent CBMs for the concept ``wing pattern.'' Columns 2-4 show the saliency map for each discrete value of concept: solid, spotted, stripe. The last column shows the saliency for the entire ``Wing pattern'' concept by averaging the saliency maps from columns 2-4. Both CBMs attend to the entire bird instead and not just to the wing. The saliency maps are computed using Integrated Gradients with Gaussian Noise baseline.}
    \label{fig:saliency}
\end{figure*}

\begin{figure*}[htb]
    \centering
    \includegraphics[width=0.5\textwidth]{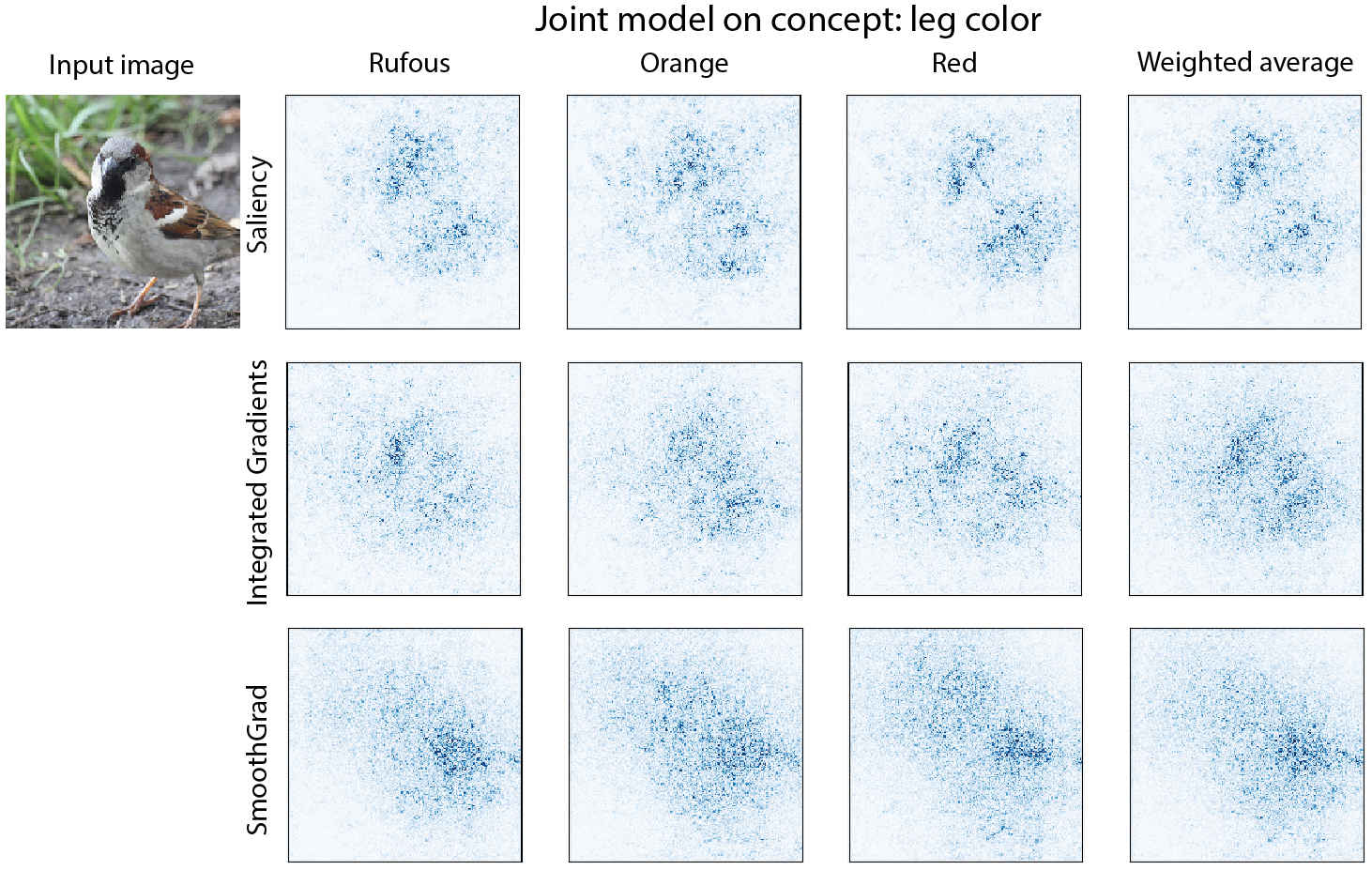}
    \caption{Joint model ``leg color.'' The bird's leg is not attended to by any saliency method.}
    \label{jnt}
\end{figure*}

\begin{figure*}[t!]
    \centering
    \includegraphics[width=0.5\textwidth]{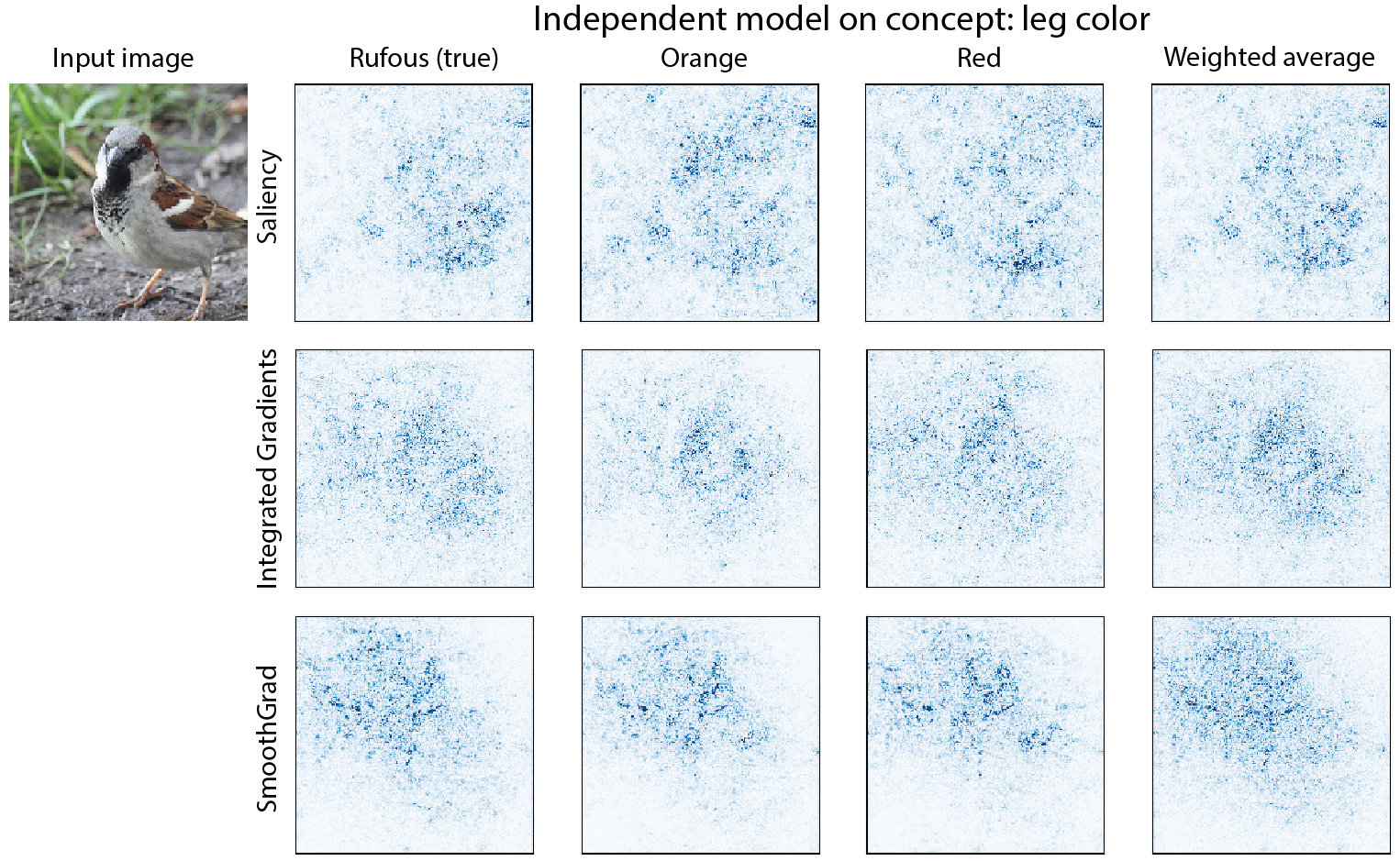}
    \caption{Independent model ``leg color.'' The bird's leg is not attended to by any saliency method.}
     \label{ind}
\end{figure*}

A desirable property of concepts is to depend on the relevant parts of the input space. For example, the wing concept ought to correspond to the wing of a bird in input space. Several works suggest that models can assign importance to spurious regions of input space: \citet{ribeiro2016should} showed that in husky versus wolf classification, their model picked up the snow in the background as important since husky images in the training data were more likely to contain snow. We identify regions in input space to which the CBM concepts attend.
To investigate this, we construct saliency maps from the concept layer to the input space. Figure \ref{fig:saliency} illustrates  saliency maps for joint and independent bottlenecks. In both cases, the wing concept attends to the entire bird. In Figure~\ref{jnt} and Figure~\ref{ind}, we find that similarly the ``leg color'' concept does not attend to the leg of the bird for multiple saliency methods.
We believe that existing saliency methods map be ill-equipped to study attribution for concept bottlenecks. Future work can attempt to develop methods to right this. 
Appendix \ref{appendix:additional_posthoc} provides more results and compares different post hoc methods. Note that if the concepts are correctly predicted, then one can still argue that we can get some understanding of which concepts are important by looking at the intervenability of the concept layer.

\section{Discussion}
In this work, we study concept bottleneck models (CBMs). Our results suggest that CBMs may not learn as intended. We call into question a CBM's ability to meet three desiderata: interpretability, predictability, and intervenability. While the joint CBM learns information about the target before the concept layer, we find that the independent CBM may be the only variant that achieves all three desiderata given current methods for CBMs. Using existing saliency map methods, we find that none of the concepts learned to map semantically meaningful representations in input space. 
We hope that future work attempts to show concepts to domain experts: were concepts to map to something meaningful in input space, such methods could be used to discover concepts that are important to targets but have not been pre-specified by experts.

\clearpage
\subsubsection*{Acknowledgments}
The authors thank Pang Wei Koh, Been Kim, and Percy Liang for their helpful comments.
AM acknowledges support from the Cambridge ESRC Doctoral Training Partnership. MA acknowledges support from the George and Lilian
Schiff Foundation. UB acknowledges support from DeepMind and the Leverhulme Trust via the Leverhulme Centre for the Future of Intelligence (CFI) and from the Mozilla Foundation. AW acknowledges support from a Turing AI Fellowship under grant EP/V025379/1, The Alan Turing Institute under EPSRC grant EP/N510129/1 and TU/B/000074, and the Leverhulme Trust via CFI.

\bibliographystyle{plainnat}
\bibliography{main}

\clearpage
\appendix

\section{Experimental Set-Up}
\label{setup}

\subsection{Datasets and Hyperparameters}
For the x-grading task, we use $36,369$ x-rays from the Osteoarthritis Initiative (OAI) to predict the Kellgren-Lawrence grade (KLG), a common scale used by radiologists to measure the severity of osteoarthritis~\citep{peterfy2008osteoarthritis}; like~\citet{koh2020concept}, we aim to predict KLG from ten ordinal, clinically relevant concepts: these concepts are instance-specific, which means that examples with the same $\*y$ can have different $\*c$. Our ten concepts can be used by medical practitioners to assess KLG~\citep{kellgren1957radiological}. 

For the bird identification task, we use $11,788$ photographs of birds to predict the correct bird species from 200 possible options: this dataset is known as CUB~\citep{wah2011caltech}. We use the 112 concepts used from~\citep{koh2020concept}. These concepts are class-level, which means that all training data of the same class $\*y$ have the same $\*c$. The concepts for CUB are also one-hot encoded, which means the concept ``wing color'' is transformed from a categorical to a one hot (``has\_wing\_color\_red,''  ``has\_wing\_color\_brown,''  etc.).

When training a CBM, we admit that our results will highly depend on the $\lambda$ we use to weight our concept loss: the larger, the more enforced our expert-specified concepts are. For CUB, we use $\lambda=0.01$. For OAI, we use $\lambda=1$.


\subsection{Post Hoc Interpretability}
We leverage post hoc interpretability tools to analyze where feature space our CBMs attend. We use Integrated Gradients~\citep{sundararajan2017axiomatic} and SmoothGrad~\citep{smilkov2017smoothgrad} to obtain saliency maps of which features are important to a model's predictions. Since we have two models ($g$ from inputs to concepts and $f$ from concepts to targets), we get two types of saliency maps. First, we  identify which concepts are important to a specific target by applying post hoc interpretability techniques to $f$, i.e., we can confirm if an American Goldfinch has a black wing. We also identify where concepts lie in input space by applying the same techniques to $g$, i.e., we can check if the tibia concept corresponds to the tibia in the x-ray itself. 

When obtaining saliency maps from $g$, we have to overcome the one-hot-nature of our concepts. We obtain a separate saliency map for each ``beak color'' with the CUB data. While it may be elucidating to show concepts for each color, we may want a summary saliency map for ``beak color.'' We use two potential methods to overcome this hurdle: (1) we take the mean of the saliency maps for each of the categories of a given concept; or (2) we take a weighted average of the saliency maps, wherein we weigh the saliency map by the softmax of the one-hot-encoded concepts themselves. We leverage out-of-box implementations for all saliency map experiments~\citep{kokhlikyan2020captum}.

Appendix \ref{appendix:additional_posthoc} shows additional post hoc interpretability results.

\clearpage
\section{Additional post hoc results}
\label{appendix:additional_posthoc}
Additional post hoc results using three saliency methods: Saliency (or Gradient), Integrated Gradients~\citep{sundararajan2017axiomatic} and Smoothgrad~\citep{smilkov2017smoothgrad}. All saliency maps are computed on images from the training set on which both the Joint and Independent models predicts correctly the shown concepts. The saliency maps of Concept Bottleneck Models consistently focus on areas outside the concept of interest.


\begin{figure*}[hb]
    \centering
    \includegraphics[width=0.85\textwidth]{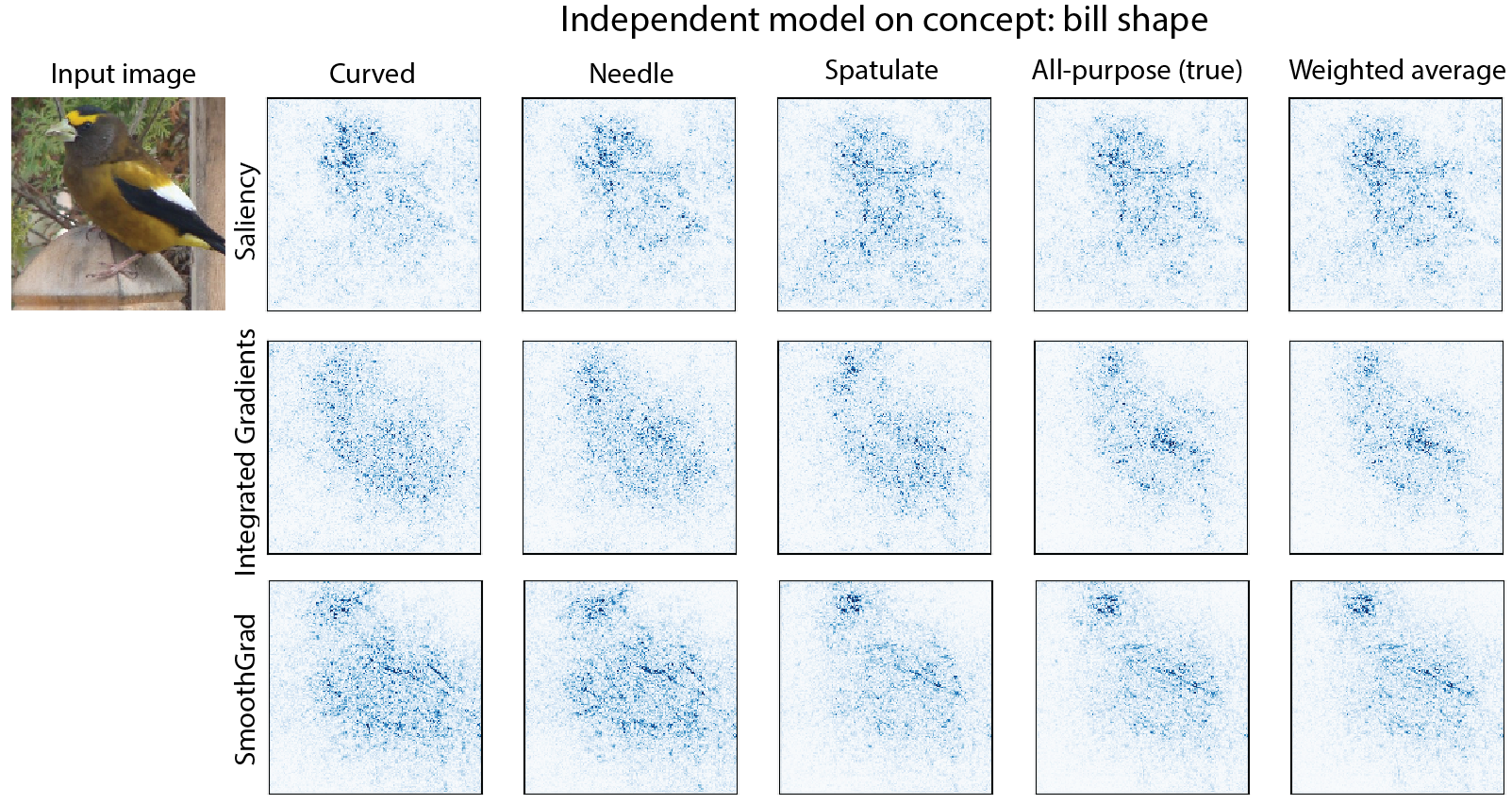}
    \caption{Independent model ``bill shape'' concept.}
\end{figure*}
\begin{figure*}[hb]
    \centering
    \includegraphics[width=0.85\textwidth]{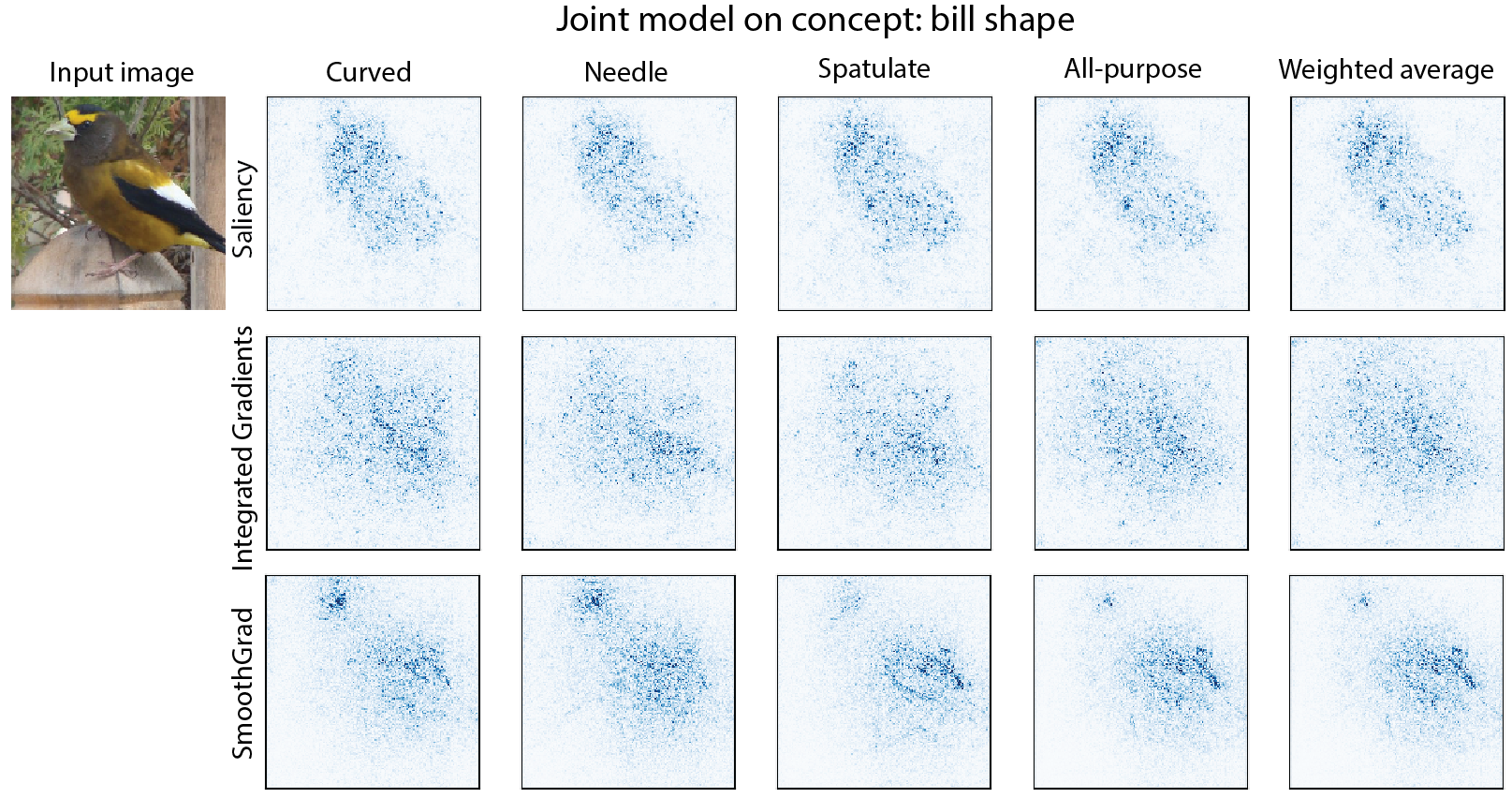}
    \caption{Joint model ``bill shape'' concept.}
\end{figure*}
\begin{figure*}[t]
    \centering
    \includegraphics[width=0.85\textwidth]{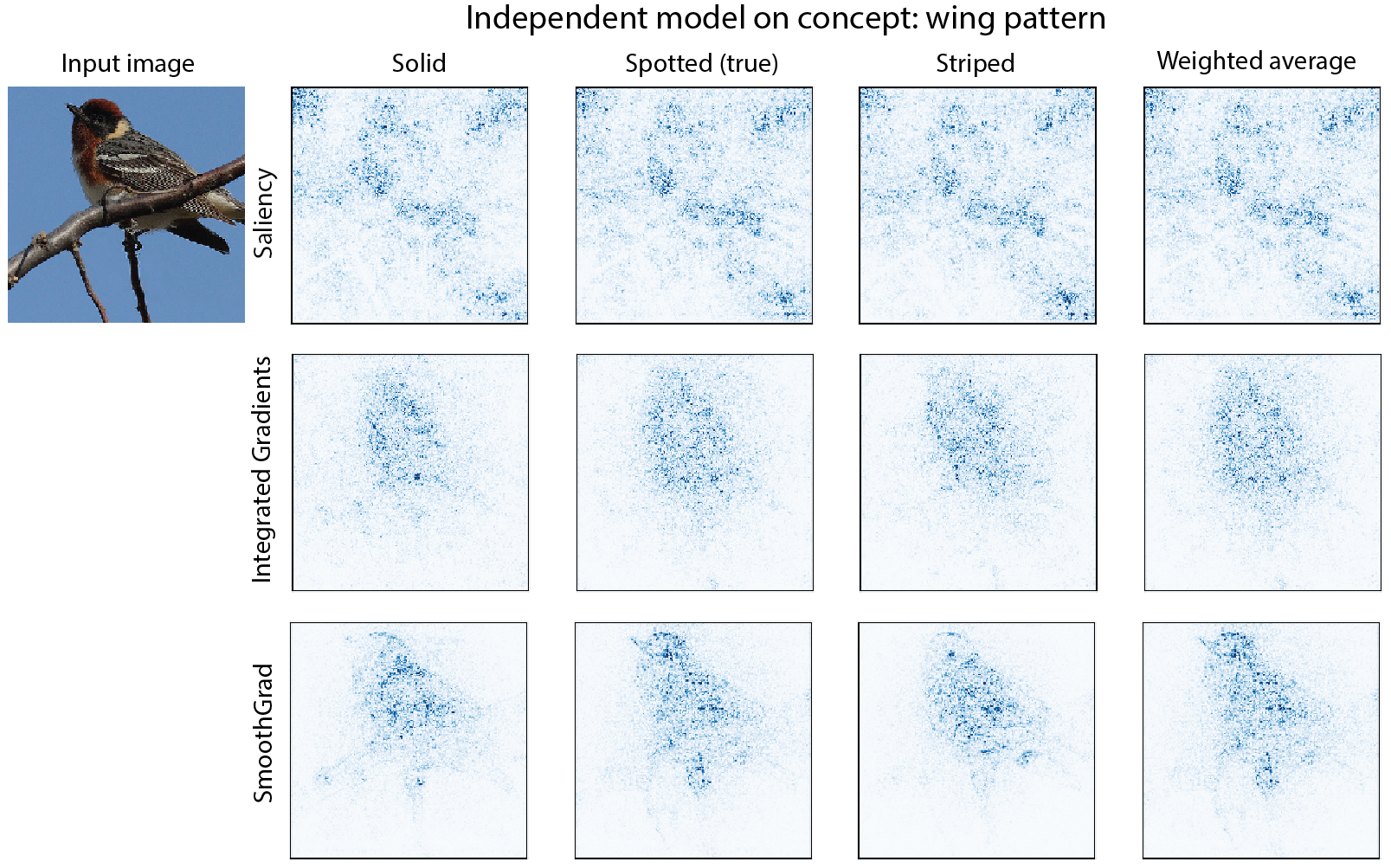}
    \caption{Independent model ``wing pattern'' concept.}
\end{figure*}


\begin{figure*}
    \centering
    \includegraphics[width=0.85\textwidth]{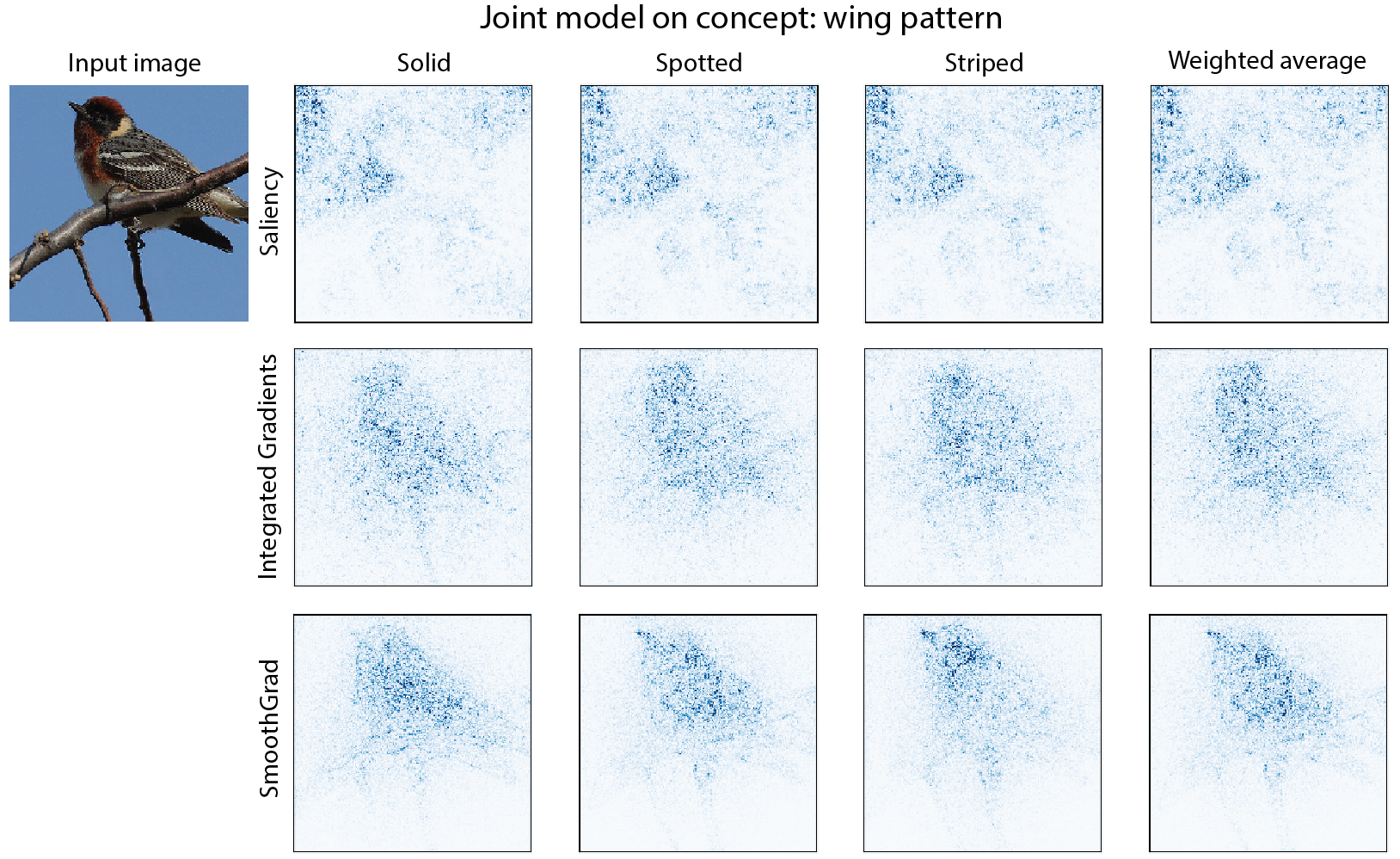}
    \caption{Joint model ``wing pattern'' concept.}
\end{figure*}

\end{document}